\documentclass[runningheads]{llncs}
\usepackage{graphicx}
\usepackage{amsmath,amssymb} 
\usepackage{color}
\usepackage{cite}
\usepackage{booktabs}
\usepackage{multirow}
\usepackage{algorithm, algpseudocode}
\usepackage[misc]{ifsym}
\usepackage{relsize}
\usepackage[hypertexnames=false,colorlinks=true,allcolors=blue]{hyperref}
\usepackage{autonum}
\usepackage{color}

\newcommand{\miniparagraph}[1]{\vspace{0.5em}\noindent\textsf{\textbf{#1:~}}}

\newcommand{\figref}[1]{Fig.\,\ref{#1}}
\newcommand{\tabref}[1]{Table\,\ref{#1}}

\renewcommand{\eqref}[1]{Eq.\,\ref{#1}}
\newcommand{\mathbar}{\text{-}}

\LetLtxMacro\latexincludegraphics\includegraphics
\renewcommand{\includegraphics}[2][]{\latexincludegraphics[#1]{#2}}

\algdef{SE}[SUBALG]{Indent}{EndIndent}{}{\algorithmicend\ }%
\algtext*{Indent}
\algtext*{EndIndent}

\makeatletter
\newcommand{\algmargin}{\the\ALG@thistlm}
\makeatother

\algnewcommand{\parState}[1]{\State%
    \parbox[t]{\dimexpr\linewidth-\algmargin}{\strut #1\strut}}

\begin{document}
\pagestyle{headings}

\mainmatter

\title{FSNet: An Identity-Aware Generative Model for Image-based Face Swapping} 

\titlerunning{FSNet: An Identity-Aware Generative Model for Image-based Face Swapping}
\authorrunning{R. Natsume et al.}
\author{Ryota Natsume\inst{1} \and Tatsuya Yatagawa\inst{1} \and Shigeo Morishima\inst{1, 2} }
\institute{Waseda University, 3-4-1, Ohkubo, Shinjuku, Tokyo, 169-8555, Japan \\ \email{ryota.natsume.26@gmail.com}, ~~~ \email{tatsy@acm.org} \and Waseda Research Institute of Science and Engineering, Tokyo, Japan \\ \email{shigeo@waseda.jp}}

\maketitle

\begin{abstract}
    This paper presents FSNet, a deep generative model for image-based face swapping. Traditionally, face-swapping methods are based on three-dimensional morphable models (3DMMs), and facial textures are replaced between the estimated three-dimensional (3D) geometries in two images of different individuals. However, the estimation of 3D geometries along with different lighting conditions using 3DMMs is still a difficult task. We herein represent the face region with a latent variable that is assigned with the proposed deep neural network (DNN) instead of facial textures. The proposed DNN synthesizes a face-swapped image using the latent variable of the face region and another image of the non-face region. The proposed method is not required to fit to the 3DMM; additionally, it performs face swapping only by feeding two face images to the proposed network. Consequently, our DNN-based face swapping performs better than previous approaches for challenging inputs with different face orientations and lighting conditions. Through several experiments, we demonstrated that the proposed method performs face swapping in a more stable manner than the state-of-the-art method, and that its results are compatible with the method thereof.
    
\end{abstract}

\section{Introduction}
\label{sec:introduction}

\begin{figure}[tb]
    \centering
    \includegraphics[width=\linewidth]{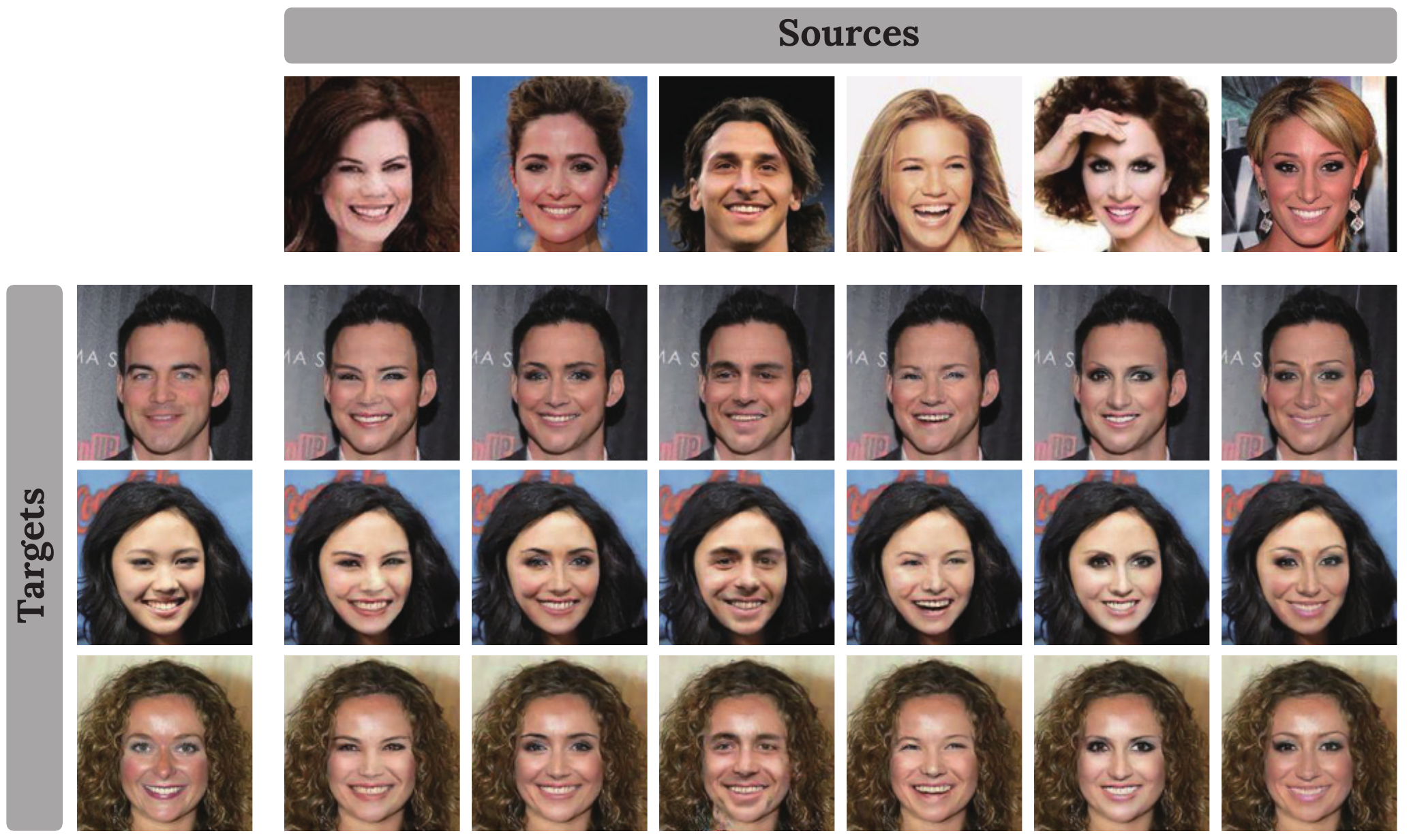}
    \caption{Results of image-based face swapping using our method. In this figure, face regions of the target images in the left column are replaced with the faces of the source images in the top row. The supplementary document provides additional results for other input faces.}
    \label{fig:face-swap}
\end{figure}

Face image editing has become increasingly prevalent owing to the growth of social networking services and photo-retouching software. To respond to the potential demand for creating more attractive face images with such photo-retouching software, many studies have been introduced for a range of applications including face image analysis~\cite{blanz02,cao14,liu15,zhang16_mtcnn} and manipulation~\cite{blanz04,bitouk08,yang11,chai12,kemelmacher16} in computer vision and graphics communities. In such applications, face swapping is an important technique owing to its broad applications such as photomontage~\cite{blanz04}, virtual hairstyle fitting~\cite{kemelmacher16}, privacy protection~\cite{bitouk08,mosaddegh14,korshunova16}, and data augmentation for machine learning~\cite{hassner13,mclaughlin15,masi16}. 

As its name indicates, face swapping replaces the face region of a target image with that in a source image. Traditional approaches~\cite{blanz04,nirkin17} uses three-dimensional morphable models (3DMMs) to estimate face geometries and their corresponding textures. Subsequently, the textures of the source and target images are swapped using the estimated texture coordinates. Finally, the replaced face textures are re-rendered using the lighting condition estimated from the target image. However, these methods are prone to fail in estimating the face geometries or lighting conditions in practice. The inaccurate estimations typically cause noticeable artifacts because human eyes can well detect slight mismatches of these geometries and lighting conditions.

In contrast to the methods above, several recent studies have applied deep neural networks (DNNs) to face swapping. Bao et al.~\cite{bao17} proposed a conditional image generation technique using a DNN, known as CVAE-GAN. They performed face swapping by considering face identities as image conditioners. A similar technique was used in ``FakeApp''~\cite{fakeapp}, an easy-to-use application software for image-based face swapping using a DNN. Korshunova et al.~\cite{korshunova16} considered face identities as artistic styles in neural style transfer~\cite{gatys16}, and performed face swapping by fine-tuning the pre-trained network using a dozens of images of an individual. Although these approaches facilitated the use of deep-learning techniques for face swapping, they share a common problem in that the users must prepare multiple input images of an individual. Meanwhile, Natsume et al.~\cite{natsume18} proposed a DNN for face image editing that only uses a single source image and a single target image. Even though their method can be applied to a wide range of applications including face swapping, the hair regions of their face-swapped results are no longer the same as those in the original images.

To address the problems above, we propose FSNet, a novel DNN for image-based face swapping. The proposed method disentangles face appearance as a latent variable that is independent of the face geometry and the appearance of the non-face region, including hairstyles and backgrounds. The latent variables for the two face appearances of two input images are swapped, and are combined with the latent variables for the non-face parts of the counterpart images. Once the network of FSNet is trained using a large-scale face image dataset, FSNet does not require any additional fine-tuning and performs face swapping only with a single source image and a single target image. As shown in \figref{fig:face-swap}, the faces are swapped appropriately in that face identities in the source images are preserved well and composed naturally with the non-face regions of the target images. In this study, we evaluated the face-swapping results using a number of image assessment metrics and demonstrated that FSNet achieves more stable face swapping than the state-of-the-art methods. In addition to the stability, the quality of the FSNet results is compatible to the methods thereof. The technical contributions of FSNet are summarized as follows:
\begin{enumerate}
    \item It is a new DNN for image-based face swapping that uses only a single source and a single target images, and does not require any additional fine-tuning.
    \item While face swapping, it well preserves both the face identity in a source image and the appearances of hairstyle and background region in a target image.
    \item It performs high-quality face swapping even for typical challenging inputs with different face orientations and with different lighting conditions.
\end{enumerate}

\section{Related Work}
\label{sec:related-work}

Face swapping has been studied for a range of applications including photomontage~\cite{blanz04}, virtual hairstyle fitting~\cite{kemelmacher16}, privacy protection~\cite{bitouk08,mosaddegh14,korshunova16}, and data augmentation for large-scale machine learning~\cite{masi16}. Several studies~\cite{yang11,mosaddegh14} have replaced only parts of the face, such as eyes, nose, and mouth between images rather than swapping the entire face. A popular approach for face swapping is based on the 3DMM~\cite{blanz04,nirkin17}. Fitting a 3DMM to a target face yields the face geometry, texture map, and lighting condition ~\cite{blanz02,cao14}. A face-swapped appearance is generated by the replacement of the face textures and the subsequent re-rendering of the face appearance using the estimated lighting condition.

The primary drawback of these approaches is the difficulty in the accurate estimation of three-dimensional (3D) face geometries and lighting conditions from single images. The failure estimation often causes noticeable visual artifacts. To alleviate this problem, Bitouk et al.~\cite{bitouk08} proposed an image-based face swapping without the 3DMM. To avoid the estimation of face geometries, they leveraged a large-scale face database. Their system searches a target image whose layout is similar to that of a source image. Subsequently, these face regions of two similar images are swapped using boundary-aware image composition. A more sophisticated approach was recently proposed by Kemelmacher-Shlizerman~\cite{kemelmacher16}. She carefully designed a handmade feature vector to represent face image appearances and improved the accuracy of searching similar faces successfully. However, these methods do not allow the users to choose both the source and target images; therefore, they are not applicable to arbitrary face image pairs.

Several recent studies have applied deep neural networks for image-based face swapping. Bao et al.~\cite{bao17} indicated that their conditional image generation technique can alter face identities by conditioning the generated images with an identity vector. Meanwhile, Korshunova et al.~\cite{korshunova16} applied the neural style transfer~\cite{gatys16} for face swapping by considering the face identities as the artistic styles in the original style transfer. However, these recent approaches still have a problem. They require at least dozens of images of an individual person to obtain a face-swapped image. Collecting that many images is possible, albeit unreasonable for most non-celebrities. 

Another recent study~\cite{bao18} proposed an identity-preserving GAN for transferring image appearances between two face images. While the purpose of this study is close to that of face swapping, it does not preserve the appearances of non-face regions including hairstyles and backgrounds. Several studies for DNN-based image completion~\cite{iizuka17,chen18} have presented demonstrations of face appearance manipulation by filling the parts of an input image with their DNNs. However, the users can hardly estimate the results of these approaches because they only fill the regions specified by the users such that the completed results imitate the images in the training data.

\section{FSNet: A Generative Model for Face Swapping}
\label{sec:rsgan}

\begin{figure*}[tb]
	\centering
	\includegraphics[width=\linewidth]{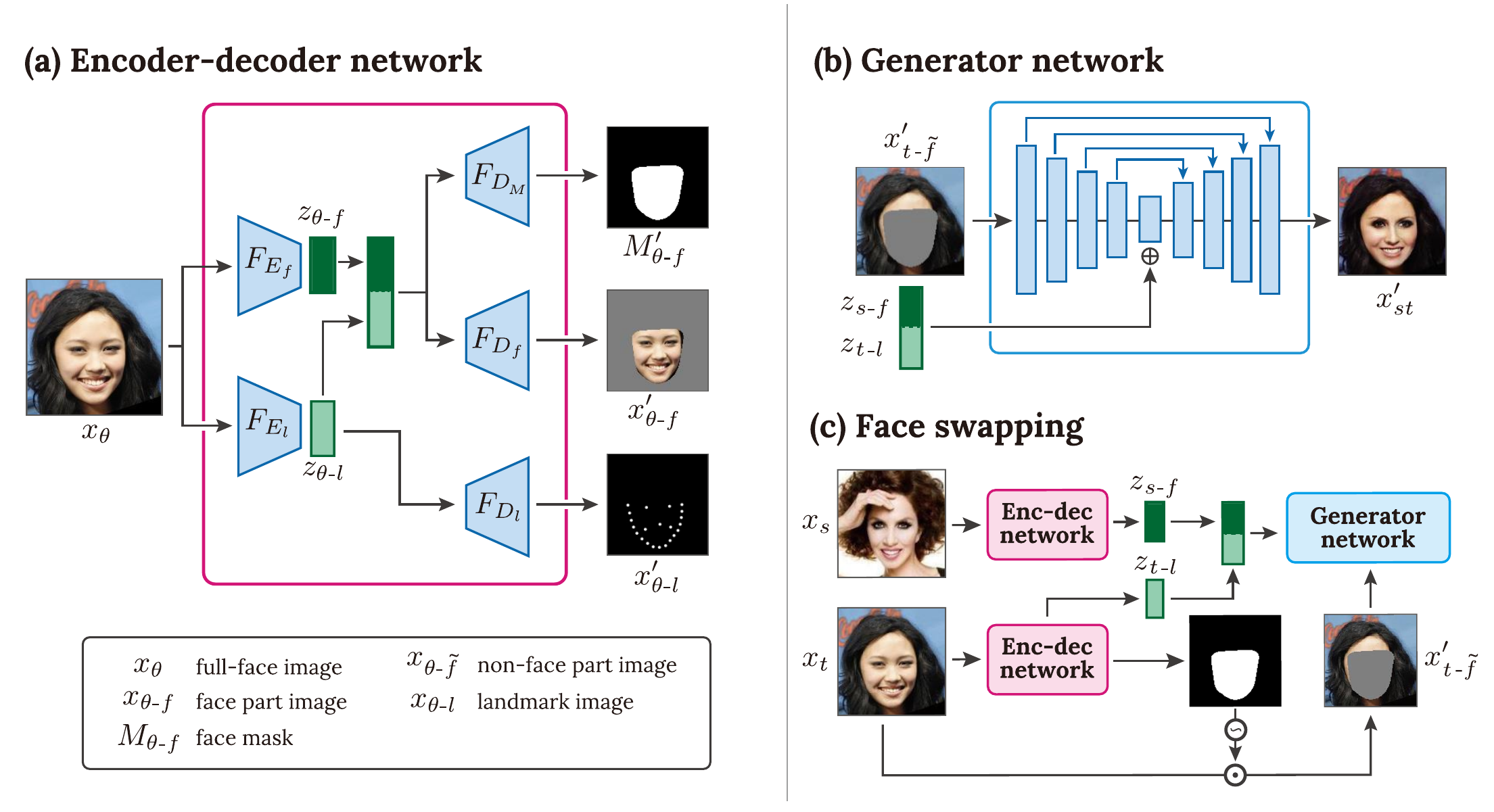}
	\caption{Network architecture of FSNet. The network consists of two partial networks, i.e., (a) encoder-decoder network, and (b) generator network. The encoder-decoder network obtains a latent variable for a face appearance that is independent of the face geometry and appearance of the non-face part. The generator network synthesizes a face-swapped result from the latent variable and non-face part of another image.}
	\label{fig:network}
\end{figure*}

The architecture of FSNet is shown in \figref{fig:network}. The network is separated into two parts and each of them performs one of two different tasks. The first part, i.e., encoder-decoder network in \figref{fig:network}(a), disentangles a face appearance as a latent variable from a source image. The architecture of this part is based on the variational autoencoder~\cite{kingma13} (VAE), and the latent variable can be obtained from the middle layer of the network. The second part, i.e., generator network in \figref{fig:network}(b), synthesizes a new face part such that it fits the non-face part of a target image. The architecture of this part is based on the U-Net~\cite{ronneberger15} and it synthesizes the face part by concatenating the latent variable with the feature map provided in the middle layer of the network. In the following subsections, we first elaborate the two partial networks;  subsequently, we describe an image dataset for training the networks. In addition, we provide the detailed network architecture of FSNet in the supplementary document.　The notations used in this paper are also summarized in the supplementary document.

\subsection{Encoder-decoder network}
\label{ssec:enc-dec-network}

The training of FSNet requires sampling the full face image $x_{\theta}$ and three intermediate images as in \figref{fig:network}. Here, $\theta \in \{ s, t\}$, and $x_s$ and $x_t$ represent a source and target image in face swapping, respectively. The three intermediate images, i.e., face mask $M_{\theta \mathbar f}$, face part image $x_{\theta \mathbar f}$, and landmark image $x_{\theta \mathbar l}$, will be compared to the outputs from the encoder-decoder network.

As shown in \figref{fig:network}(a), the encoder-decoder network outputs face mask $M'_{\theta \mathbar f}$, face part image $x'_{\theta \mathbar f}$, landmark image $x'_{\theta \mathbar l}$, and non-face part image $x'_{\theta \mathbar \tilde{f}}$. In the encoder-decoder network, the full face image $x_{\theta}$ is first encoded by two different encoders $F_{E_f}$ and $F_{E_l}$. Following the standard VAEs, these encoders output the means and standard deviations of the corresponding standard normal distributions. Subsequently, the latent variables $z_{\theta \mathbar f}$ and $z_{\theta \mathbar l}$ are sampled from the following distributions:
\begin{align}
z_{\theta \mathbar f} = \mathcal{N} \left( \mu_{\theta \mathbar f}, \sigma_{\theta \mathbar f} \right), &\quad \left( \mu_{\theta \mathbar f}, \sigma_{\theta \mathbar f}^2 \right) = F_{E_f}(x_{\theta}), \label{eq:latent-space-sampling} \\
z_{\theta \mathbar l} = \mathcal{N} \left( \mu_{\theta \mathbar l}, \sigma_{\theta \mathbar l} \right), &\quad \left( \mu_{\theta \mathbar l}, \sigma_{\theta \mathbar l}^2 \right) = F_{E_l}(x_{\theta}),
\end{align}
where $\mu_{\theta}$ and $\sigma_{\theta}^2$ are the mean and variance of $z_{\theta}$. The three decoders $F_{D_M}$, $F_{D_f}$, and $F_{D_l}$ reconstruct face mask $M'_{\theta \mathbar f}$, face image $x'_{\theta \mathbar f}$, and landmark image $x'_{\theta \mathbar l}$, respectively:
\begin{align}
M'_{\theta \mathbar f} = F_{D_M}(z_{\theta　\mathbar f}, z_{\theta \mathbar l}), \quad x'_{\theta \mathbar f} = F_{D_f}(z_{\theta \mathbar f}, z_{\theta \mathbar l}), \quad x'_{\theta \mathbar l} = F_{D_l}(z_{\theta \mathbar l}).
\end{align}
To encode only geometry-independent information in $z_{\theta \mathbar f}$, we input both $z_{\theta \mathbar f}$ and $z_{\theta \mathbar l}$ to $F_{D_f}$ and $F_{D_M}$, similarly for $F_{E_f}$. 

\subsection{Generator network}
\label{ssec:generator-network}

The architecture of generator network $G$ is based on the U-Net~\cite{ronneberger15}, as shown in \figref{fig:network}(b). Unlike the original U-Net, the generator network receives latent variables and concatenates them with a feature map given in the middle layer of the network. Moreover, the generator network receives the non-face part image $x'_{t \mathbar \tilde{f}} = x'_{t \mathbar f} \odot \tilde{M}_{t \mathbar f}^{\prime}$  rather than the full-face image $x_{t}$. Here, $\odot$ denotes pixel-wise multiplication and $\tilde{M}$ denotes an inversion of the mask $M$. To stabilize the training process, we add Gaussian noises of  standard deviation $\sigma = 0.05$ to the non-face part image when the network is trained. In the middle layer of the generator network, latent variables $z_{s \mathbar f}$ and $z_{t \mathbar l}$ are tiled and concatenated with the feature map. Subsequently, the concatenated feature map is fed to the latter part of the U-Net structure. Finally, we can obtain a face-swapped image $x'_{st}$ as an output of the generator network. We denote the operation of the generator network as follows:
\begin{equation}
x'_{st} = G(x_{t \mathbar \tilde{f}}, z_{s \mathbar f}, z_{t \mathbar l}).
\end{equation}
When the two same images $x_{s}$ and $x_{t}$ ($s = t$) are input to the proposed network, the generator network should reproduce the input full-face image $x_{s}$ ($= x_t$). We denote the reconstructed image as $x_s^{\prime}$. It is noteworthy that the masked image $x_{t \mathbar \tilde{f}}^{\prime}$, which is one of the inputs of the generator network, is computed by the encoder-decoder network only using the full-face image $x_t$. Therefore, face swapping can be performed with only the source and target image themselves, and the user need not prepare intermediate images used in the training. 

\subsection{Training}
\label{ssec:training}

The proposed network is trained similarly as VAE-GAN~\cite{larsen_vaegan}. In other words, the two partial networks are trained with VAE objectives and GAN objectives, separately. In addition, the proposed network is also trained with an identity loss to preserve the face identities in the face swapping results. We define the identity loss using the triplet loss~\cite{cheng16_triplet}. Therefore, we sample a triplet of images consisting of anchor sample $x_{s_1}$, positive sample $x_{s_2}$, and negative sample $x_t$. The anchor and positive samples are used as source images, and the negative sample is used as a target image. The different identities in these three images are ignored when we evaluate the VAE and GAN objectives.

\miniparagraph{VAE objectives}
For three outputs $M_{\theta \mathbar f}^{\prime}$, $x_{\theta \mathbar f}^{\prime}$, and $x_{\theta \mathbar l}^{\prime}$,  we define the reconstruction losses using the corresponding ground truth images in the training data. We evaluate the cross entropy losses for the face masks and landmark images and an L1 loss for the face part images:
\begin{align}
\mathcal{L}_{\theta \mathbar M}^{rec} &= \mathbb{E} [ L_{CE} (M_{\theta \mathbar f}, M'_{\theta \mathbar f})], \\
\mathcal{L}_{\theta \mathbar f}^{rec} &= \mathbb{E} [ \|x_{\theta \mathbar f} - x'_{\theta \mathbar f} \|_1 ], \\
\mathcal{L}_{\theta \mathbar l}^{rec} &= \mathbb{E} [ L_{CE} (x_{\theta \mathbar l}, x'_{\theta \mathbar l})], 
\end{align}
where $L_{CE}$ denotes a function for the cross entropy loss. In addition, we define another reconstruction loss $\mathcal{L}_{\theta}^{rec}$ between the full-face image $x_{\theta}$ and its corresponding reconstruction $x'_{\theta}$ as well. For imposing more reconstruction loss for the pixels in the foreground region, i.e., the face and hair regions, we define the loss with the foreground mask $M_{\theta \mathbar FG}$ as follows:
\begin{equation}
\mathcal{L}_{\theta}^{rec} = \mathbb{E} \Big[ \big\| (x_\theta - x'_\theta) \odot (M_{\theta \mathbar FG} + \beta (1 - M_{\theta \mathbar FG})) \big\|_1 \Big].
\end{equation}
In our implementation, we used the parameter $\beta = 0.5$ to halve the losses in the background. To evaluate the means and standard deviations given with the encoders, we employed a latent classifier as in $\alpha$-GAN~\cite{rosca17} rather than evaluating the Kullback-Leibler loss in the standard VAEs. Let $C_{\omega}$ be the latent classifier and $z \sim \mathcal{N}(0, 1)$ be a random vector sampled with the standard normal distribution. Therefore, we can define the latent classification loss as follows:
\begin{equation}
\mathcal{L}_{\theta \mathbar f}^{lat} = -\mathbb{E} [\log C_{\omega} (z_{\theta \mathbar f})] -\mathbb{E} [\log (1 - C_{\omega} (z_{\theta \mathbar f}))].
\label{eq:latent-loss}
\end{equation}
Equally, $\mathcal{L}_{\theta \mathbar M}^{lat}$ and $\mathcal{L}_{\theta \mathbar l}^{lat}$ are defined for $z_{\theta \mathbar M}$ and $z_{\theta \mathbar l}$.

\miniparagraph{GAN objectives}
As the standard GAN, both the encoder-decoder and generator networks are trained adversarially with several discriminators. To evaluate the real and synthesized images, we used two discriminators, i.e., global discriminator $D_g$ and patch discriminator $D_p$. The global discriminator distinguishes whether an image is a real sample or a synthesized image. The patch discriminator, which is originally introduced as a part of PatchGAN~\cite{isola17}, distinguishes whether a local patch of the image is from a real sample or a synthesized image. In addition to $x_{\theta}$ and $x'_{\theta}$, we also synthesize images with a random face using normal random vectors instead of $z_{\theta \mathbar f}$ and $z_{\theta \mathbar l}$. Let $\hat{x}'_{\theta}$ be such a random face image, we define global and patch adversarial losses $\mathcal{L}_{\theta}^{adv \mathbar g}$ and $\mathcal{L}_{\theta}^{adv \mathbar p}$ as follows:
\begin{align}
\mathcal{L}_{\theta}^{adv \mathbar \{ g, p \}} =& - \mathbb{E} \left[ \log D_{ \{ g, p \} } (x_{\theta}) \right] \nonumber \\
& - \mathbb{E} \left[ \log (1\! - \! D_{ \{ g, p \} } (x'_{\theta})) \right] \nonumber \\
& - \mathbb{E} \left[ \log (1\! - \! D_{ \{ g, p \} } (\hat{x}'_{\theta})) \right]. \label{eq:adv-loss}
\end{align}

\miniparagraph{Identity loss}
In the CelebA dataset, which we used in the experiments, identity labels are assigned to all the images. A straightforward method to evaluate the identity of a synthesized image is to train an identity classifier. However, human faces are typically similar between two different people. We found that training by this straightforward approach is unstable and can be easily stuck in a local minimum. Alternatively, we employed the triplet loss~\cite{cheng16_triplet} to evaluate the similarity of identities in two face images. The triplet loss is defined for a triple of image samples, i.e., anchor, positive, and negative samples. These samples are first encoded to the feature vectors; subsequently, the distances of the feature vectors are computed for the anchor and positive samples, and the anchor and negative samples. The triplet loss is defined to broaden the difference between these distances. To train the network, we generate two face-swapped results $x'_{s_1 t}$ and $x'_{s_2 t}$ from three input images $x_{s_1}$, $x_{s_2}$, and $x_t$. The triplet losses are defined for the triplets $\{ x_{s_1}, x_{s_2}, x_{t} \}$, $\{ x'_{s_1 t}, x_{s_1}, x_t)\}$, and $\{ x'_{s_2 t}, x_{s_2}, x_t \}$, where images in each triplet denote \{anchor, positve, negative\} samples. For a triplet $\{ x_{s_1}, x_{s_2}, x_t \}$, the triplet loss is defined using a feature extractor $F_{E_{id}}$ as in the original study~\cite{cheng16_triplet}:
\begin{align}
\mathcal{L}_{\{ s_1, s_2, t \}}^{id}
&= \min \left( 0, \| F_{E_{id}} (x_{s_1}) - F_{E_{id}} (x_{s_2}) \|_2^2 + \alpha_1 - \| F_{E_{id}} (x_{s_1}) - F_{E_{id}} (x_{t}) \|_2^2 \right) \\
&+ \alpha_3 \min \left(0,  \| F_{E_{id}} (x_{s_1}) - F_{E_{id}} (x_{s_2}) \|_2^2 - \alpha_2 \right).
\end{align}
In our implementation, we set the parameters as $\alpha_1 \! = \! 1.0$, $\alpha_2 \!=\! 0.1$, and $\alpha_3 \!=\! 0.5$. To normalize the color balances in the face images, we subtract the average pixel colors from the image and divide it by the standard deviation of the pixel colors before feeding the images to $F_{E_{id}}$.

\miniparagraph{Hyper parameters}
The overall loss function for training FSNet is defined by a weighted sum of the loss functions above:
\begin{align}
\mathcal{L} &= \lambda_{f}^{rec} \mathcal{L}_{f}^{rec} + \lambda_{M}^{rec} \mathcal{L}_{M}^{rec} + \lambda_{l}^{rec} \mathcal{L}_{l}^{rec} \\
& + \lambda^{lat} (\mathcal{L}_{f}^{lat} + \mathcal{L}_{M}^{lat} + \mathcal{L}_{l}^{lat}) \nonumber \\
& + \lambda^{adv \mathbar g} \mathcal{L}^{adv \mathbar g} + \lambda^{adv \mathbar p} \mathcal{L}^{adv \mathbar p} \\
& + \lambda^{id} \mathcal{L}^{id}
\end{align}
In this equation, we simply wrote $\mathcal{L}$ as an average of the corresponding losses for $s_1$, $s_2$, and $t$ for VAE and GAN objectives, and $\mathcal{L}^{id}$ as the sum of all the triplet losses for the identity losses. 
We empirically determined the weighting factors as $\lambda_{f}^{rec} = \lambda_{M}^{rec} = 4{,}000$, $\lambda_{l}^{rec} = 2{,}000$, $\lambda^{lat} = 30$, $\lambda^{adv \mathbar g} = 20$, $\lambda^{adv \mathbar p} = 30$, and $\lambda^{id} = 100$. In our experiment, the loss functions were minimized by mini-batch training using the ADAM optimizer~\cite{kingma14_adam} with an initial learning rate of 0.0002, $\beta_1 = 0.5$, and $\beta_2 = 0.999$. A mini-batch includes 20 images for each of $x_{s_1}$, $x_{s_2}$, and $x_{t}$; therefore, the size of the mini-batch was 60.

\begin{figure}[tb]
    \centering
    \includegraphics[width=\linewidth]{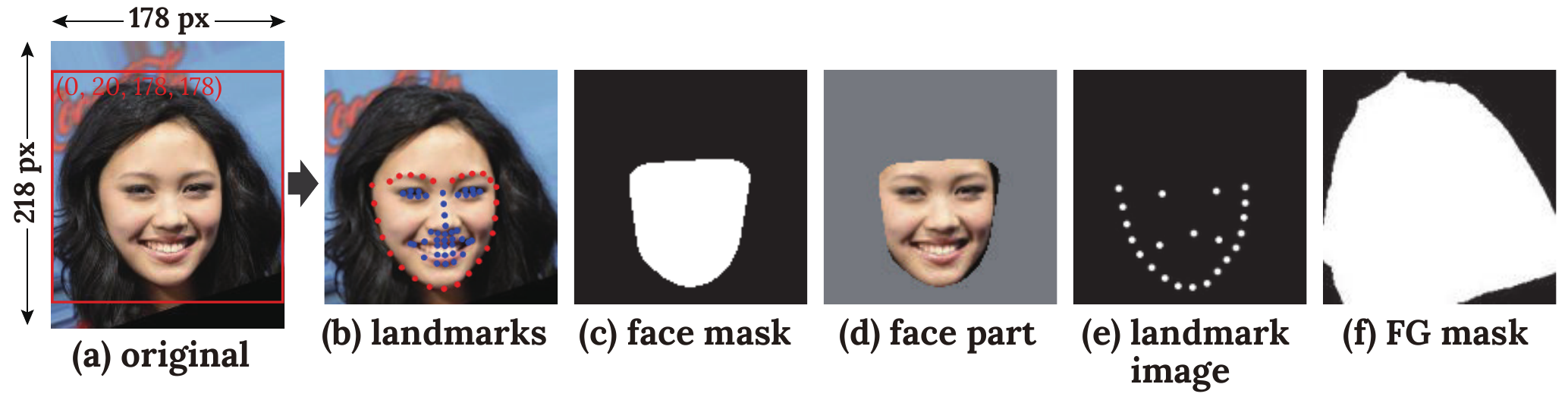}
    \caption{Dataset preparation from each image in CelebA~\cite{liu15}. To reduce the size of the dataset, we include only (a), (c), (e), and (f) in the dataset, and other intermediate images are computed while training FSNet.}
    \label{fig:dataset}
\end{figure}

\subsection{Datasets}
\label{ssec:datasets}

The dataset for training FSNet includes four types of images, i.e., the original full-face image (\figref{fig:dataset}(a)), face mask image (\figref{fig:dataset}(c)), landmark image (\figref{fig:dataset}(e)), and foreground mask (\figref{fig:dataset}(f)). All these images are generated computationally for each image in CelebA.

For an original full-face image, we first extract 68 facial landmarks (\figref{fig:dataset}(b)) with Dlib~\cite{king09}, which is a typically used machine-learning library. A convex hull is computed from the 41 landmarks that correspond to the eyes, nose, and mouth, which are indicated with blue circles in \figref{fig:dataset}(b). The convex hull is stretched to obtain a face mask (\figref{fig:dataset}(c)). The hull is stretched 1.3 times along the horizontal direction and 1.4 times along the vertical direction. 

Subsequently, we dilate the mask by 3\% of the image width to ensure that the mask boarders are slightly inside the face contours and include the eyebrows inside the mask. The face part image (\figref{fig:dataset}(d)) is obtained by applying the face mask to the input image. We use landmarks on the eyes instead of the eyebrows because the eyebrows are often hidden by bangs. Compared to the eyebrows, the landmarks on the eyes are less likely to be hidden, and the face masks can be more appropriately defined with them. 

The landmark image \figref{fig:dataset}(e) includes 5 landmarks inside the face region and 17 landmarks on the face contour. The top two internal landmarks correspond to the eye center positions, and are calculated by averaging the position of the eye landmarks. That in the middle corresponds to the nose position, and is represented by a landmark on the tip of nose. The two bottom ones correspond to the mouse position and are represented by two landmarks on two ends of the mouse. The 17 contour landmarks are represented by those on the face contour among the original 68 landmarks. These 22 landmarks are splatted on the landmark image as circles with a radius of 3\% of the image width. Finally, The foreground mask (\figref{fig:dataset}(f)) is detected using a state-of-the-art semantic segmentation method, PSPNet~\cite{zhao16_pspnet}. Pixels labeled as ``person'' are used as the foreground mask.

All of these images are cropped from $178 \times 218$ pixel region of the original image by a square of $178 \times 178$ pixels whose top-left corner is at $(0, 20)$. We used the cropped images after resizing them to $128 \times 128$. While processing images in CelebA, we could extract facial landmarks properly for 195,361 images out of 202,599 images. Among these 195,361 images, we used 180,000 images for training and the other 15,361 images for testing.

\section{Results}
\label{sec:results}

\begin{figure}[tb]
	\centering
	\includegraphics[width=\linewidth]{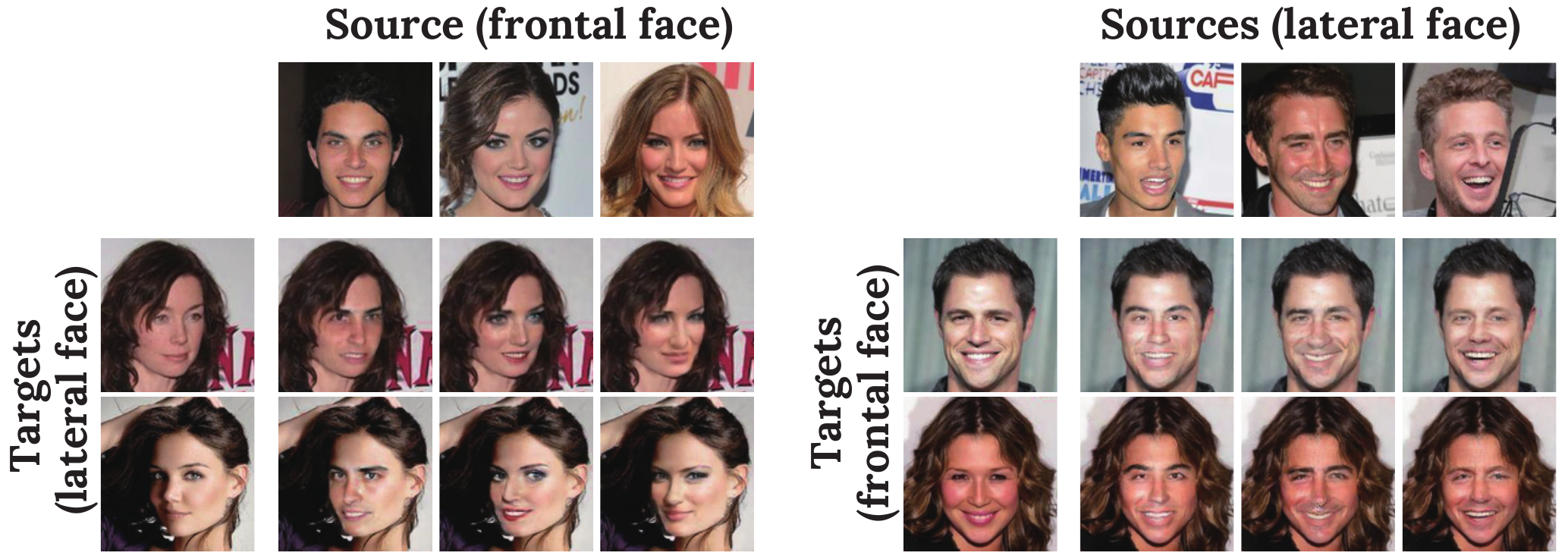}
	\caption{Face swapping between images with different face orientations. The left group shows the swapping results from frontal faces to lateral faces, while the right group shows those from lateral faces to frontal faces.}
	\label{fig:orientation}
\end{figure}

\begin{figure}[tb]
	\centering
	\includegraphics[width=\linewidth]{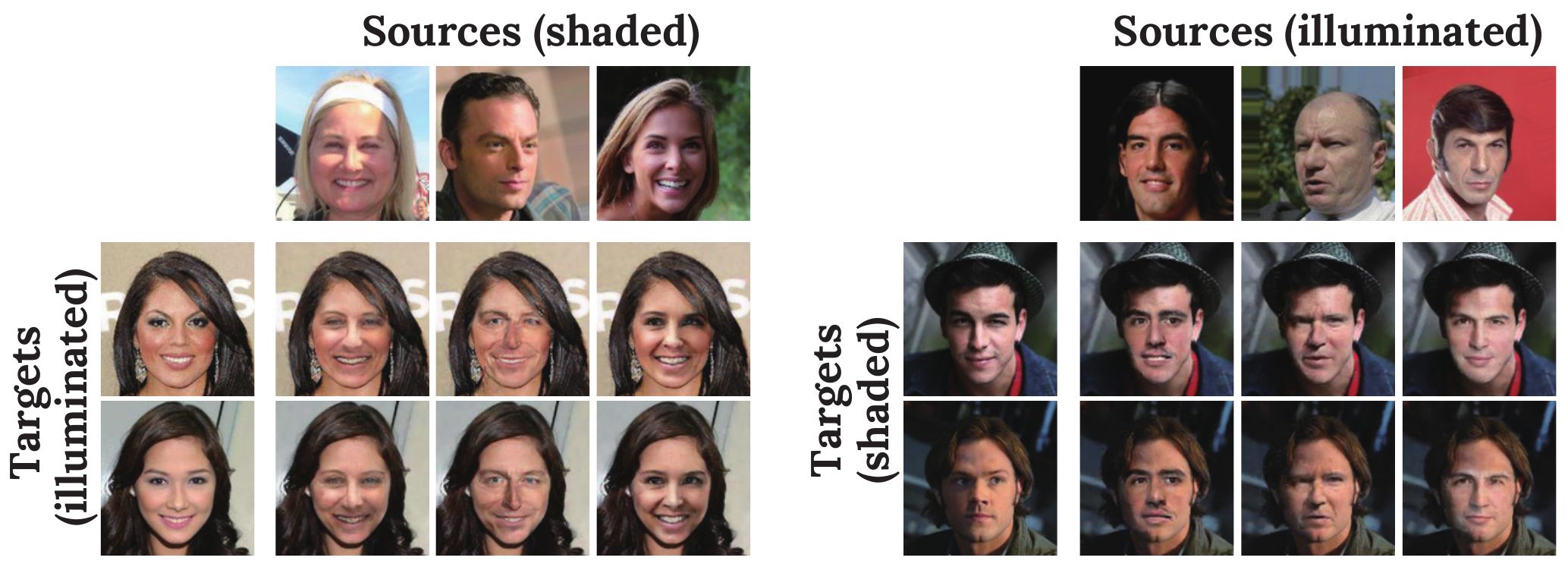}
	\caption{Face swapping between images with different lighting conditions. The left group shows the swapping results from shaded faces to uniformly illuminated faces, while the right group shows those from uniformly illuminated faces to shaded faces.}
	\label{fig:shading}
\end{figure}

\begin{figure}[tb]
	\centering
	\includegraphics[width=\linewidth]{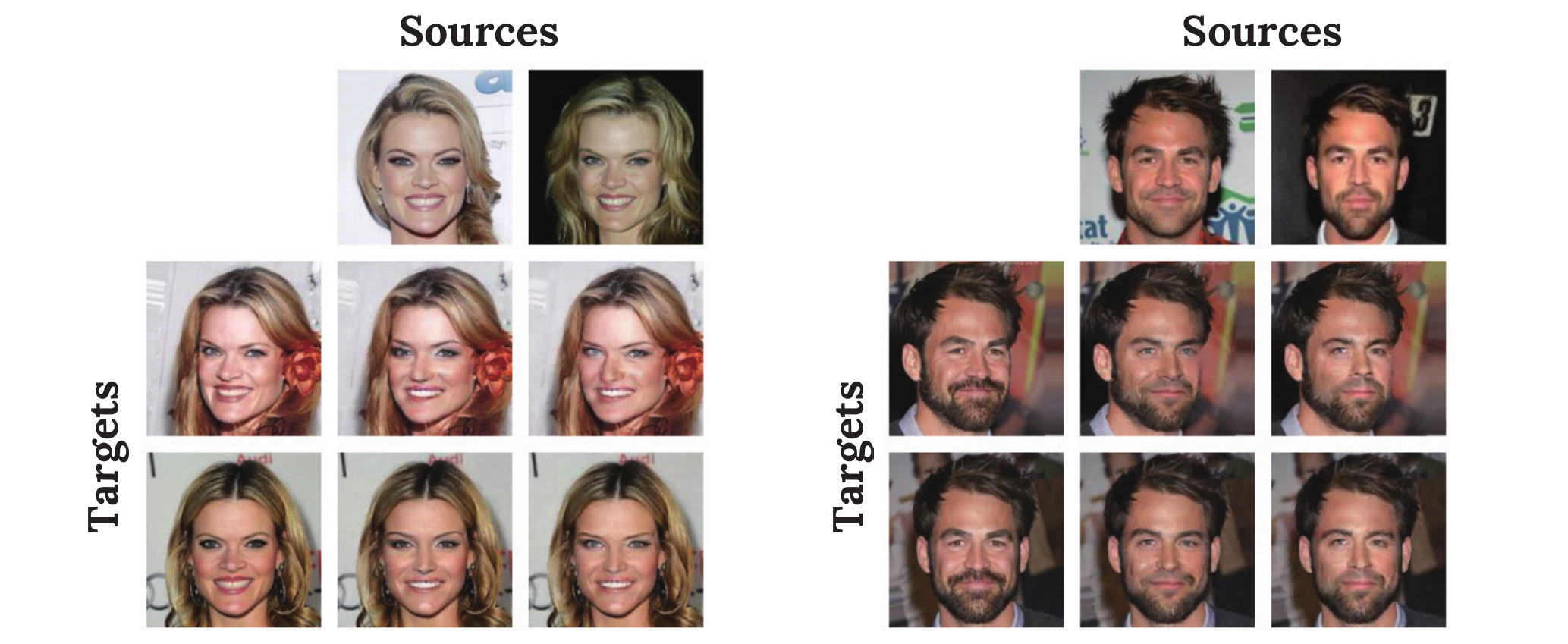}
	\caption{Face swapping between images of the same person. The four images in each group are those of an individual.}
	\label{fig:same-identity}
\end{figure}


This section presents our face swapping results for various face images. The proposed method was implemented with TensorFlow in Python, and executed on a computer with an Intel Xeon 3.6 GHz E5-1650 v4 CPU, NVIDIA GeForce GTX TITAN~X GPU, and 64 GB RAM. We trained the proposed network over 180,000 global steps. The training required approximately 72 hours using a single GPU. All the results herein were generated using test images not included in the training data.

Figure~\ref{fig:face-swap} shows the face swapping results for several face images. In this figure, the source images are in the top row and the target images are in the left column. As shown in this figure, face swapping is performed appropriately for such a variety of inputs. In addition, we tested the input images with several challenging cases to demonstrate the robustness of the proposed method. First, we swapped faces between images with different face orientations. The results are shown in \figref{fig:orientation}. In this figure, one of the sources or target images shows a frontal face and the other shows a lateral face. As shown in this figure, the face appearances including their identities are transferred appropriately to the target image even though the face orientations differed significantly. Next, we tested the images with different lighting conditions. As shown in \figref{fig:shading}, one of the sources or target images shows a uniformly illuminated face while the other shows a face lit from the side. When shaded faces are transferred to uniformly illuminated faces, the shades are removed appropriately from the faces. Furthermore, when uniformly illuminated faces are transferred to shaded faces, the overall appearances of the results are natural whereas the shades in the target images are not necessarily observed in the results. Thus, the proposed method achieves face swapping even in such challenging cases.

We also evaluated the capability of preserving face identities by swapping faces of a single individual using the proposed method. The results are illustrated in \figref{fig:same-identity}. In this figure, each group of images includes two input images for source images and the other two for target images. All of these four images show the faces of a single person. As shown in this figure, the second and third rows in each group are almost identical. These results demonstrate that the proposed network can preserve the facial identities in the input images appropriately.

\begin{table}[tb]
    \centering
    \caption{Performance evaluation in capabilities of identity preservation and image qualities.}
    \label{tab:experiments}
    \begin{tabular*}{\linewidth}{l@{\extracolsep{\fill}}lcccc}
        \toprule
        & & \multicolumn{2}{c}{Same person} & \multicolumn{2}{c}{Different people} \\
        \cmidrule{3-4} \cmidrule{5-6}
        & & Abs. errors & MS-SSIM & OpenFace & Inception score \\
        \midrule
        \multirow{2}{*}{VAE-GAN~\cite{larsen_vaegan}} & Avg. & 0.115 & 0.619 & 1.591 & 2.142 \\
                                                                                     & Std. & 0.041 & 0.105 & 0.499 & 0.137 \\
        \midrule[0.1mm]
        \multirow{2}{*}{$\alpha$-GAN~\cite{rosca17}} & Avg. & 0.099 & 0.705 & 1.353 & 2.056 \\
                                                                                    & Std. & 0.040 & 0.099 & 0.487 & 0.082 \\
        \midrule[0.1mm]
        \multirow{2}{*}{Nirkin et al.~\cite{nirkin17}} & Avg. & 0.024 & 0.956 & 0.811 & 2.993 \\
                                                                                 & Std. & 0.010 & 0.025 & 0.749 & 0.229 \\
        \midrule[0.1mm]
        \multirow{2}{*}{FSNet} & Avg. & 0.030 & 0.936 & 0.883 & 2.846 \\
                                              & Std. & 0.007 & 0.029 & 0.829 & 0.116 \\
        \midrule[0.1mm]
        \multirow{2}{*}{\parbox{27mm}{FSNet (for images Nirkin et al. failed)}} & Avg. & 0.031 & 0.933 & 0.888 & (1.235) \\
                                                                                & Std. & 0.006 & 0.025 & 0.837 & --- \\
        \bottomrule
    \end{tabular*}
\end{table}

To evaluate the proposed method quantitatively, and compare it with prior approaches, we conducted two experiments using four different metrics, as shown in \tabref{tab:experiments}. Two of these metrics are for measuring the capability of identity preservation, and the other two are for measuring the output image quality.

In these experiments, two previous studies: VAE-GAN~\cite{larsen_vaegan} and $\alpha$-GAN~\cite{rosca17} were used as baselines. Although these studies were not originally for face swapping, we performed face swapping with them in three steps. First, we compute a face mask similarly as in our dataset synthesis. Next, the face region of the source image in the mask is copy and pasted to the target image such that the two eye locations are aligned. Finally, the entire image appearance is repaired by being fed to each network. In addition to these baseline methods, we compared the proposed method with a 3DMM-based state-of-the-art method by Nirkin et al.~\cite{nirkin17}. In each experiment, we sampled 1,000 image pairs randomly.

In the first experiment, we swapped the faces between two different images of a single individual. Subsequently, we calculate the absolute difference and MS-SSIM~\cite{wang_msssim} between the input and swapped images. The results are shown in the third and fourth columns of \tabref{tab:experiments}. In the second experiment, the faces of two different people were swapped by each method. To evaluate how the identities are preserved after face swapping, we calculated a squared Euclidean distance of two feature vectors for an input and a face-swapped result using OpenFace~\cite{amos16_openface}, which is an open-source face feature extractor. In addition, we computed the inception scores~\cite{salimans16} to measure the capability of our face swapping method when applied to a broad variety of individuals. The results are shown in the fifth and sixth columns of \tabref{tab:experiments}. 

From these experiments, Nirkin et al.~\cite{nirkin17} demonstrated the best scores for all the metrics in \tabref{tab:experiments}, and the proposed method follows it closely. However, Nirkin et al.'s method sometimes fails to fit the 3DMM to one of the sources or target images, and could generate the results for only approximately 90\% of the input pairs. Meanwhile, the proposed method and the other baseline methods could generate the results for all the input pairs. Thus, the robustness of Nirkin et al.'s method can be problematic in practice. We also calculated each score with FSNet for the images whereby Nirkin et al.'s method could not generate the results. The scores in the bottom rows indicate that each of them is almost identical to that calculated with all sample images. These results demonstrate that the proposed method is advantageous because it can well generate high-quality results for arbitrary pairs of input images.

\begin{figure}[tb]
    \centering
    \includegraphics[width=\linewidth]{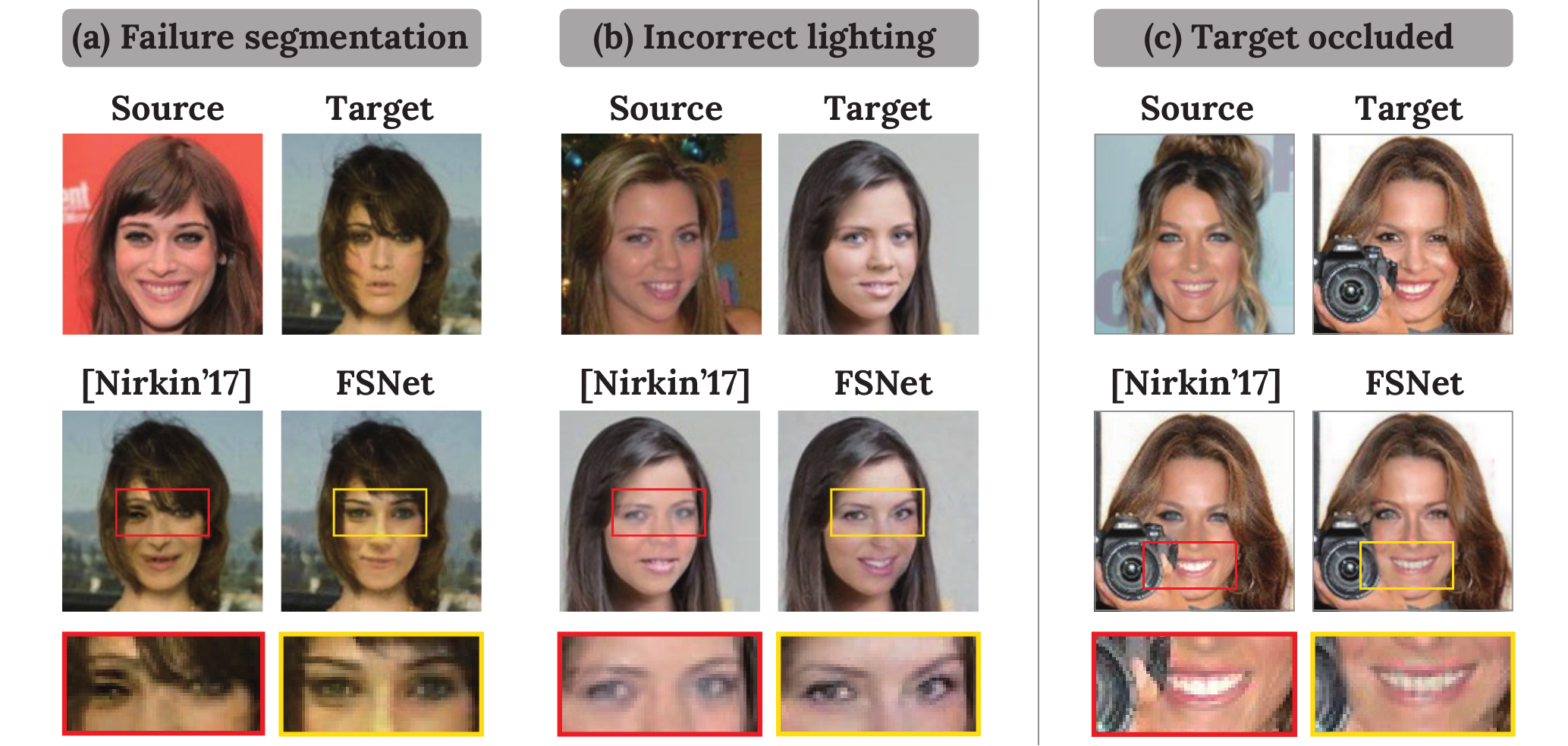}
    \caption{Typical failure cases of Nirkin et al.'s method~\cite{nirkin17} and FSNet. (a) Failure segmentation and (b) incorrect lighting estimation are those for Nirkin et al.'s method and (c) occluded target face is that for proposed FSNet.}
    \label{fig:limitations}
\end{figure}

To elaborate the comparison with the method of Nirkin et al., we present the typical failure cases of their method and our FSNet in \figref{fig:limitations}. In this figure, the first two cases exhibit the limitations of Nirkin et al.'s method, and the last one case for FSNet. First, their method and also other methods based on the 3DMM often demonstrate the inaccurate segmentation of face areas. While Nirkin et al. proposed the occlusion-aware segmentation of the face areas, its accuracy is still limited. In their results in \figref{fig:limitations}(a), the left eye of a target person remained in our face swapping result. The proposed method and other image-based method using DNNs do not require such a fragile segmentation process. Next, the estimation of lighting condition from a single image is a challenging task. The failure estimation drives the 3DMM-based methods in inappropriate face appearances. As in \figref{fig:limitations}(b), the face color in the results of Nirkin et al. was attenuated unexpectedly, and the details around eyelashes became blurry. The DNN-based methods, including FSNet, were not strongly affected by such lighting conditions, as shown in \figref{fig:shading}. Meanwhile, the proposed method cannot properly swap faces when a part of the face region is occluded. As shown in \figref{fig:limitations}(c), the camera and left hand in front of the face were lost in the face swapping result. This is because only a limited number of image samples are available for such occluded faces in the training dataset. Recently, several approaches~\cite{saito16, nirkin17} have augmented an image dataset by artificially adding random obstacles to the images, and separated face regions occluded by such obstacles successfully. Although this approach can likely be applied to the face image synthesis, we will leave it for future work owing to its uncertain applicability. It is noteworthy that the proposed method can consider a face occluded by hair as in \figref{fig:limitations}(a) because such hairstyles are included in the training data.

\section{Conclusion}
\label{sec:Conclusion}

We proposed FSNet, a deep generative model for image-based face swapping. The encoder-decoder part of the proposed network disentangles a face appearance as a latent variable that is independent of the face geometry and appearance of non-face parts. The latent variable was composed together with the non-face part of the target image, and a face-swapped image was generated by the generative network. In contrast to previous methods, our method used neither the 3DMM nor any additional fine-tuning. It performed face swapping only with a single source image and a single target image. Through a number of experiments, we demonstrated that the proposed method could perform face swapping robustly even for several challenging inputs with different face orientations and lighting conditions. In addition, the quality of the results is comparable with the state-of-the-art method~\cite{nirkin17} and performed face swapping more stably. For future work, we would like to explore its applicability to movies by introducing temporal coherency in image generation.

\section*{Acknowledgments}

This study was granted in part by the Strategic Basic Research Program ACCEL of the  Japan Science and Technology Agency (JPMJAC1602). Tatsuya Yatagawa was supported by the Research Fellowship for Young Researchers of Japan's Society for the Promotion of Science (16J02280). Shigeo Morishima was supported by a Grant-in-Aid from Waseda Institute of Advanced Science and Engineering. The authors would also like to acknowledge NVIDIA Corporation for providing their GPUs in the academic GPU Grant Program.

\bibliographystyle{splncs}
\bibliography{0919}


\clearpage
\newpage

\appendix

\renewcommand\thetable{A\thesection\arabic{table}}
\renewcommand\thefigure{A\thesection\arabic{figure}} 
\renewcommand\thealgorithm{A\thesection\arabic{algorithm}}

\setcounter{page}{1}

\title{Supplementary Document: \\FSNet: An Identity-Aware Generative Model for Image-based Face Swapping} 

\titlerunning{FSNet: An Identity-Aware Generative Model for Image-based Face Swapping}
\authorrunning{R. Natsume et al.}
\author{Ryota Natsume$^{1(\text{\Letter})}$, Tatsuya Yatagawa$^{2}$, and Shigeo Morishima$^3$}
\institute{Graduate School of Advanced Science and Engineering,\\Waseda University, Tokyo, Japan\\ $^1$\email{ryota.natsume.26@gmail.com}, $^2$\email{tatsy@acm.org}, $^3$\email{shigeo@waseda.jp}}

\maketitle

\begin{table}[h]
	\centering
	\caption{The notations used in the paper.}
	\label{tab:notations}
	\begin{tabular*}{\linewidth}{l@{\extracolsep{\fill}}l}
		\toprule
		Symbol & Meaning \\
		\midrule
		$\theta$ & input image type $s$ (source) or $t$ (target) \\
		$x_\theta$ & image of type $\theta$ \\
		$M_{\theta \mathbar f}$ & mask image for the face region of $x_\theta$ \\
		$\tilde{M}_{\theta \mathbar f}$ & mask image for the non-face region of $x_\theta$ \\
		$x_{\theta \mathbar f}$ & face part image which only shows the face part of $x_\theta$ \\
		$x_{\theta \mathbar \tilde{f}}$ & non-face part image which hides the face part of $x_\theta$ \\ 
		$x_{\theta \mathbar l}$ & landmark image which indicates the landmark positions on $x_\theta$ \\
		$M_{\theta \mathbar FG}$ & mask image for foreground region, i.e., face, hair, and body regions \\
		$x', M'$ & image and mask synthesized by neural networks \\
		$F_{E_f}, F_{E_l}$ & encoder networks for the face region and landmarks, respectively \\
		$F_{D_M}, F_{D_f}, F_{D_l}$ & decoder networks for face mask, face region, and landmarks \\
		$G$ & generator network \\
		$D_g, D_p$ & global and patch discriminator networks \\
		$C_\omega$& classifier network for latent variables, defined as in $\alpha$-GAN \cite{rosca17} \\
		$\mu, \sigma$ & mean and standard deviation for a corresponding latent variable \\
		$z$ & latent variable for a corresponding input image \\
		$\mathcal{L}_{\theta}^{rec}$ & reconstruction losses for corresponding inputs and image types \\
		$\mathcal{L}_{\theta}^{lat}$ & latent classification loss, defined as in $\alpha$-GAN \cite{rosca17} \\ 
		$\mathcal{L}_{\theta}^{adv \mathbar g}, \mathcal{L}_{\theta}^{adv \mathbar p}$ & discrimination losses for global and patch discriminators \\
		$\mathcal{L}_{s_1, s_2, t}^{id}$ & triplet loss for an image triplet $x_{s_1}$, $x_{s_2}$ and $x_t$ \\
		\bottomrule
	\end{tabular*}
\end{table}

\begin{figure}[h]
	\centering
	\includegraphics[width=\linewidth]{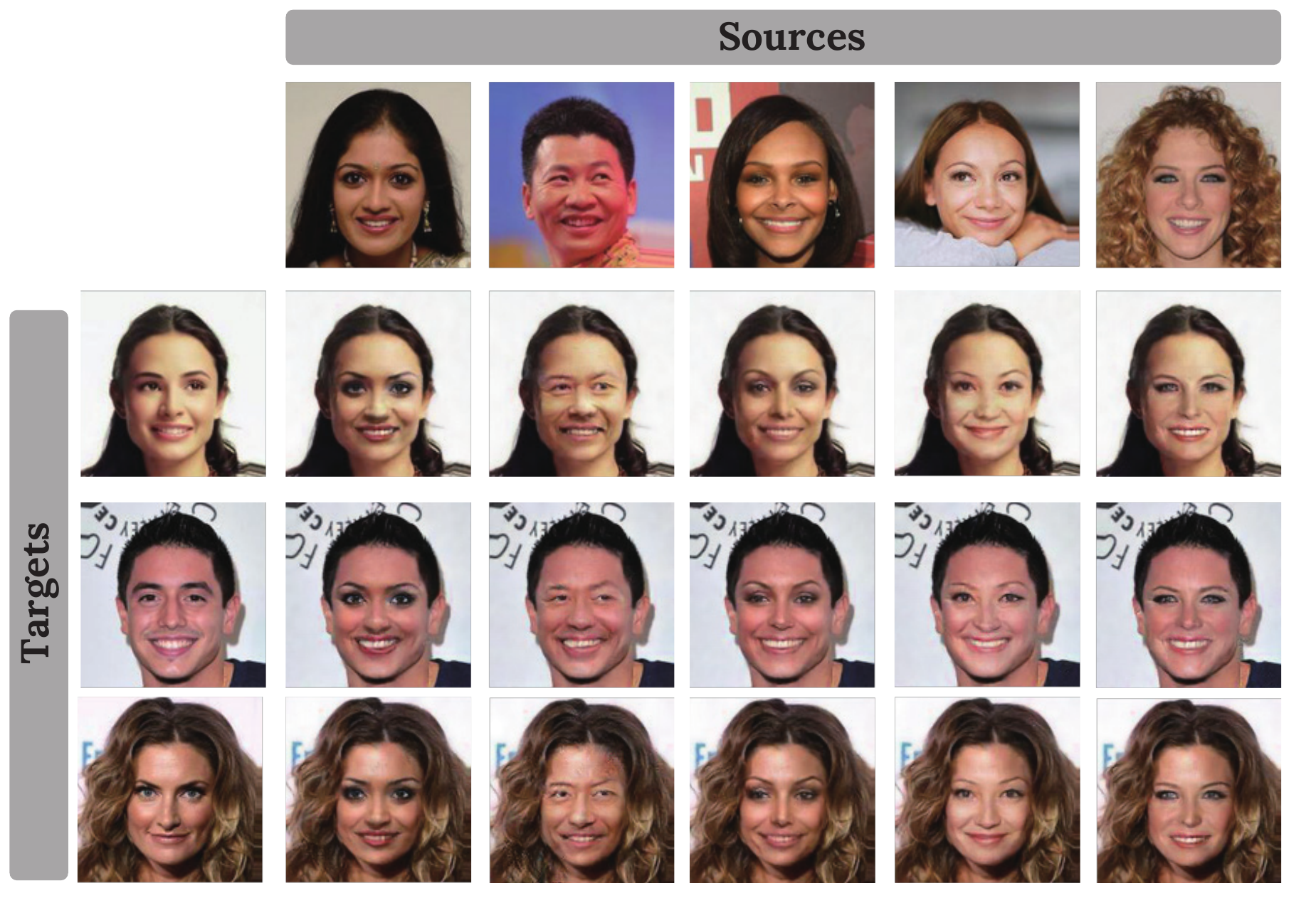}
	\caption{Additional results for face swapping.}
	\label{fig:additional-face-swap_01}
\end{figure}

\begin{figure}[h]    
	\includegraphics[width=\linewidth]{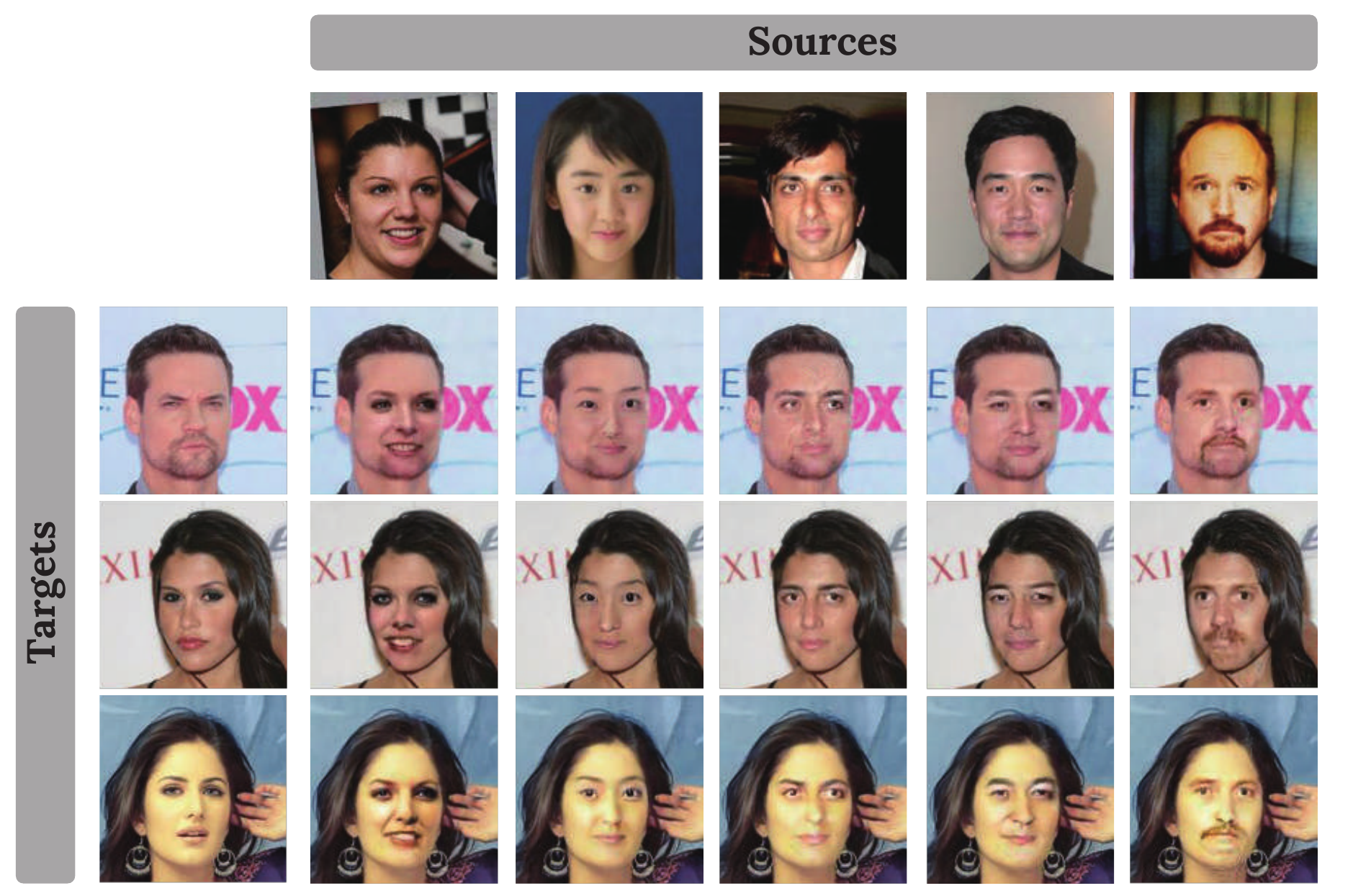} 
	\caption{Additional results for face swapping.}
	\label{fig:additional-face-swap_02}
\end{figure}

\begin{figure}[h]    
	\includegraphics[width=\linewidth]{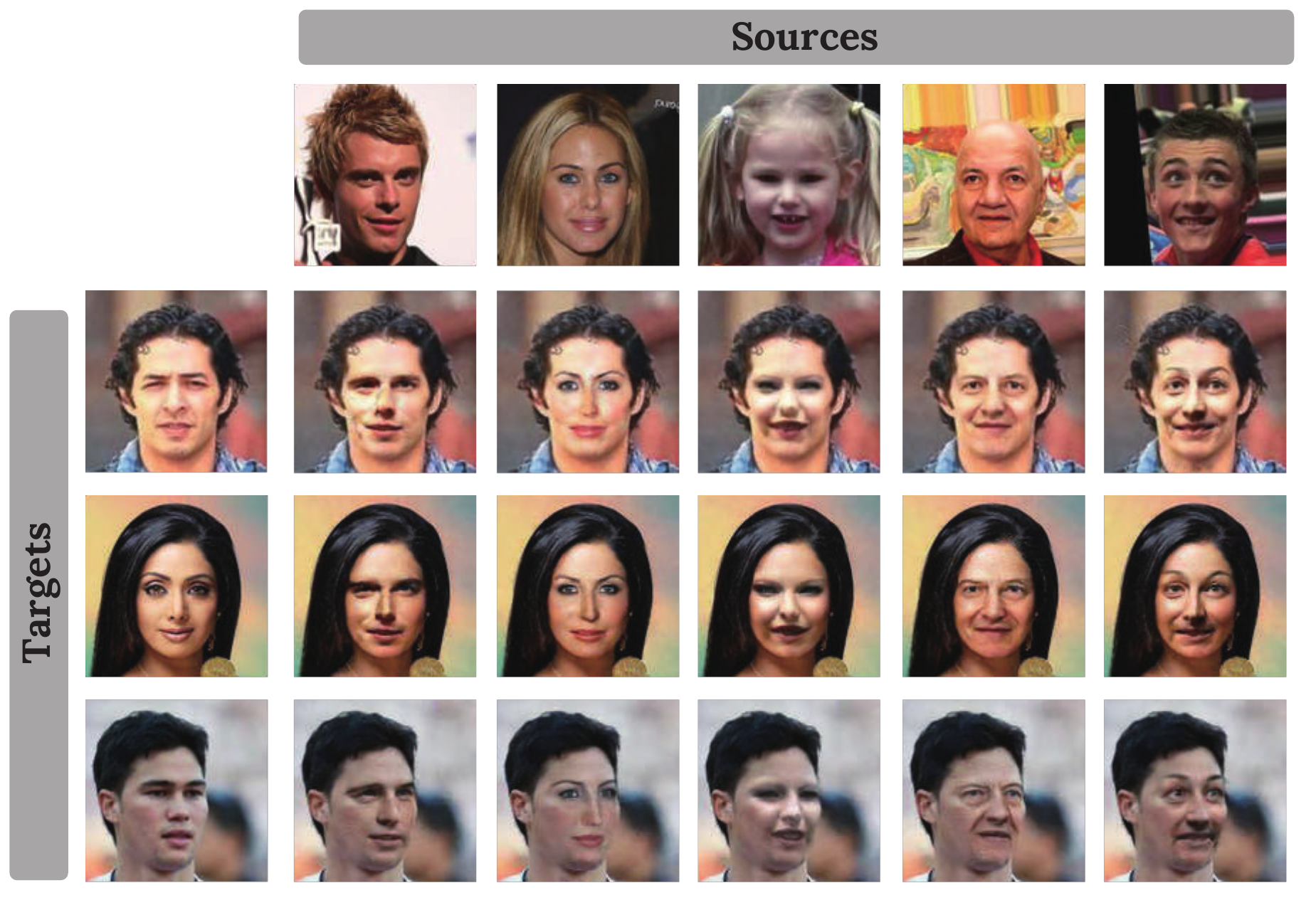} 
	\caption{Additional results for face swapping.}
	\label{fig:additional-face-swap_03}
\end{figure}

\begin{figure}[h]    
	\includegraphics[width=\linewidth]{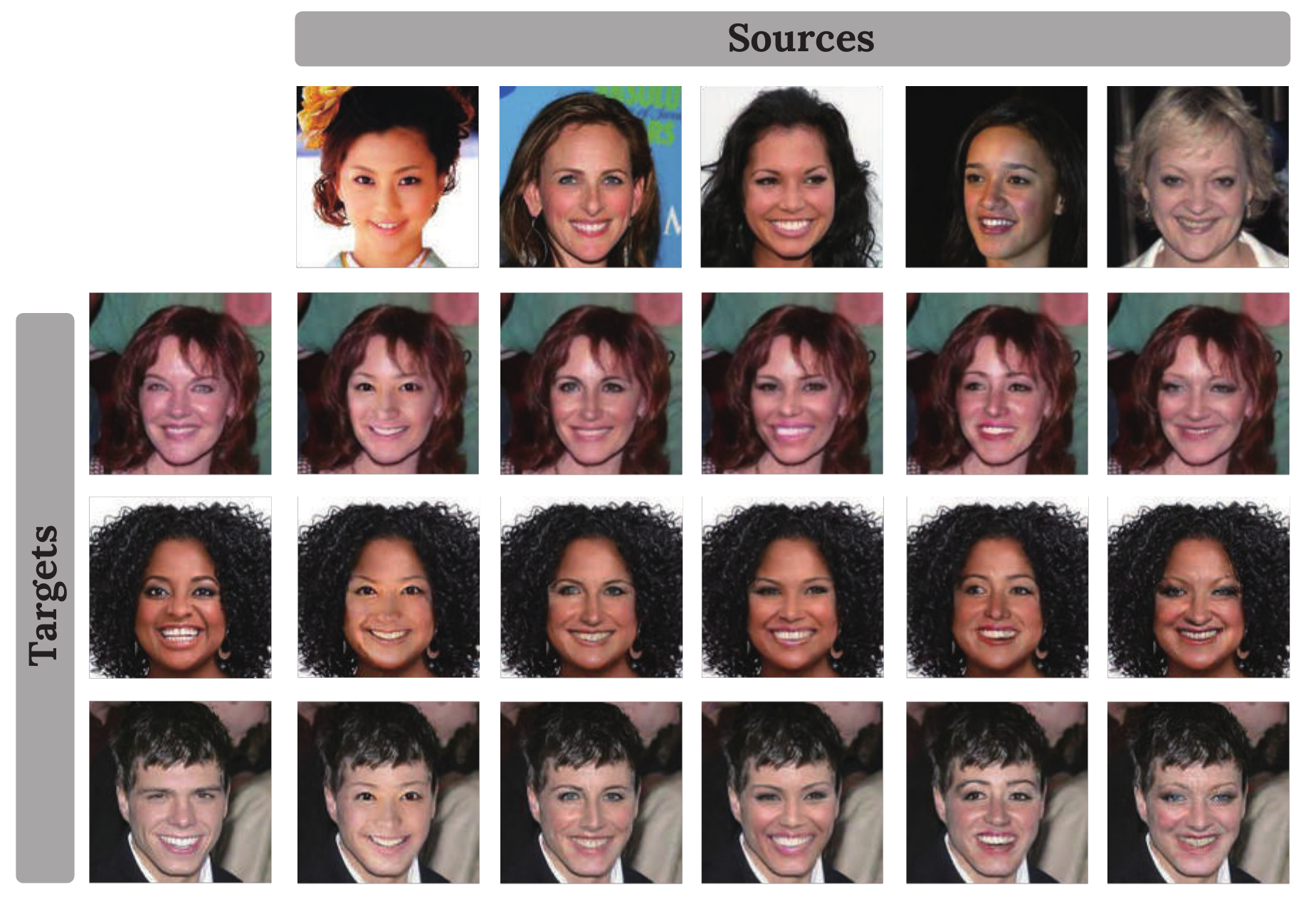} 
	\caption{Additional results for face swapping.}
	\label{fig:additional-face-swap_04}
\end{figure}

\begin{table}[h]
	\scriptsize
	\centering
	\caption{Architecture of the local discriminator $D_l(x)$.}
	\label{tab:architecture}
	\begin{tabular*}{\linewidth}{l@{\extracolsep{\fill}}llllll}
		\toprule
		Operation & Kernel  & Strides & Filters & BN & Activation & Output to \\
		\midrule
		\multicolumn{7}{l}{$D_l$ --- Input: $N_{batch} \times 128 \times 128 \times 3$. } \\
		\midrule[0.1mm]
		(1-a) Convolution & $4 \times 4$ & $2 \times 2$ & 32 & No & Leaky ReLU & (1-b) \\
		(1-b) Convolution & $4 \times 4$ & $2 \times 2$ & 64 & No & Leaky ReLU & (1-c) \\
		(1-c) Convolution & $4 \times 4$ & $2 \times 2$ & 128 & No & Leaky ReLU & (1-d) \\
		(1-d) Convolution & $4 \times 4$ & $2 \times 2$ & 256 & No & Sigmoid & --- \\
		\midrule
		\multicolumn{7}{l}{$D_g$ --- Input: $N_{batch} \times 128 \times 128 \times 3$.} \\
		\midrule[0.1mm]
		(2-a) Convolution & $4 \times 4$ & $2 \times 2$ & 32 & No & Leaky ReLU & (2-b) \\
		(2-b) Convolution & $4 \times 4$ & $2 \times 2$ & 64 & No & Leaky ReLU & (2-c) \\
		(2-c) Convolution & $4 \times 4$ & $2 \times 2$ & 128 & No & Leaky ReLU & (2-d) \\
		(2-d) Convolution & $4 \times 4$ & $2 \times 2$ & 256 & No & Leaky ReLU & (2-e) \\
		(2-e) Convolution & $4 \times 4$ & $2 \times 2$ & 512 & No & Leaky ReLU & (2-f) \\
		(2-f) Fully connected & --- & --- & 1 & No & Sigmoid & --- \\
		\midrule
		\multicolumn{7}{l}{$F_{E \mathbar x}$ --- Input: $N_{batch} \times 128 \times 128 \times 3$.} \\
		\midrule[0.1mm]
		(3-a) Convolution & $4 \times 4$ & $2 \times 2$ & 32 & No & Leaky ReLU & (3-b) \\ 
		(3-b) Convolution & $4 \times 4$ & $2 \times 2$ & 64 & No & Leaky ReLU & (3-c) \\ 
		(3-c) Convolution & $4 \times 4$ & $2 \times 2$ & 128 & No & Leaky ReLU & (3-d) \\ 
		(3-d) Convolution & $4 \times 4$ & $2 \times 2$ & 256 & No & Leaky ReLU & (3-e) \\ 
		(3-e) Convolution & $4 \times 4$ & $2 \times 2$ & 512 & No & Leaky ReLU & (3-f), (3-g) \\
		(3-f) Fully connected & --- & --- & 128 & No & Softplus & --- \\
		(3-g) Fully connected & --- & --- & 128 & No & Softplus & --- \\
		\midrule
		\multicolumn{7}{l}{$F_{E \mathbar id}$ --- Input: $N_{batch} \times 128 \times 128 \times 3$.} \\
		\midrule[0.1mm]
		(3-a) Convolution & $4 \times 4$ & $2 \times 2$ & 32 & No & Leaky ReLU & (3-b) \\ 
		(3-b) Convolution & $4 \times 4$ & $2 \times 2$ & 64 & No & Leaky ReLU & (3-c) \\ 
		(3-c) Convolution & $4 \times 4$ & $2 \times 2$ & 128 & No & Leaky ReLU & (3-d) \\ 
		(3-d) Convolution & $4 \times 4$ & $2 \times 2$ & 256 & No & Leaky ReLU & (3-e) \\ 
		(3-e) Convolution & $4 \times 4$ & $2 \times 2$ & 512 & No & Leaky ReLU & (3-f), (3-g) \\
		(3-f) Fully connected & --- & --- & 128 & No & Sigmoid & --- \\
		\midrule
		\multicolumn{7}{l}{$F_{D \mathbar x}$ --- Input: $N_{batch} \times (128 + 8 + 16) \rightarrow N_{batch} \times 1 \times 1 \times 152$.} \\
		\midrule[0.1mm]
		(5-a) Deconvolution & $4 \times 4$ & $2 \times 2$ & 512 & Yes & ReLU & (5-b) \\
		(5-b) Deconvolution & $4 \times 4$ & $2 \times 2$ & 256 & Yes & ReLU & (5-c) \\
		(5-c) Deconvolution & $4 \times 4$ & $2 \times 2$ & 128 & Yes & ReLU & (5-d) \\
		(5-d) Deconvolution & $4 \times 4$ & $2 \times 2$ & 64 & Yes & ReLU & (5-e) \\
		(5-e) Deconvolution & $4 \times 4$ & $2 \times 2$ & 32 & Yes & ReLU & (5-f) \\
		(5-f) Deconvolution & $3 \times 3$ & $1 \times 1$ & 3 & Yes & Sigmoid & --- \\
		\midrule
		\multicolumn{7}{l}{$G(x)$ --- $N_{batch} \times (128 + 8 + 128 + 8 + 16) \rightarrow N_{batch} \times 1 \times 1 \times 388$.} \\
		\midrule[0.1mm]
		(6-a) Convolution & $3 \times 3$ & $1 \times 1$ & 64 & No & ReLU & (6-b), (6-p) \\
		(6-b) Convolution & $3 \times 3$ & $2 \times 2$ & 64 & No & ReLU & (6-c) \\
		(6-c) Convolution & $3 \times 3$ & $1 \times 1$ & 128 & No & ReLU & (6-d), (6-n) \\
		(6-d) Convolution & $3\times 3$ & $2 \times 2$ & 128 & No & ReLU & (6-e) \\
		(6-e) Convolution & $3 \times 3$ & $1 \times 1$ & 256 & No & ReLU & (6-d), (6-l) \\
		(6-f) Convolution & $3\times 3$ & $2 \times 2$ & 256 & No & ReLU & (6-f) \\
		(6-g) Convolution & $3 \times 3$ & $1 \times 1$ & 512 & No & ReLU & (6-g) \\
		(6-i) Deconvolution & $3\times 3$ & $2 \times 2$ & 256 & No & ReLU & (6-j) \\
		(6-j) Convolution & $3 \times 3$ & $1 \times 1$ & 256 & No & ReLU & (6-k) \\
		(6-k) Deconvolution & $3\times 3$ & $2 \times 2$ & 128 & No & ReLU & (6-l) \\
		(6-l) Convolution & $3 \times 3$ & $1 \times 1$ & 128 & No & ReLU & (6-m) \\
		(6-m) Deconvolution & $3\times 3$ & $2 \times 2$ & 64 & No & ReLU & (6-n) \\
		(6-n) Convolution & $3 \times 3$ & $1 \times 1$ & 64 & No & ReLU & (6-o) \\
		(6-o) Deconvolution & $3\times 3$ & $2 \times 2$ & 64 & No & ReLU & (6-p) \\
		(6-p) Convolution & $3 \times 3$ & $1 \times 1$ & 3 & No & ReLU & --- \\
		\bottomrule
	\end{tabular*}
\end{table}

\clearpage

\end{document}